%% file: acl_latex.tex
\pdfoutput=1

\documentclass[11pt]{article}

\usepackage{acl}

\usepackage{times}
\usepackage{latexsym}
\usepackage{marvosym}
\usepackage{booktabs}
\usepackage{makecell}
\usepackage{graphics}
\usepackage{enumitem}
\usepackage{amsmath}
\usepackage{graphicx}
\usepackage{amssymb}
\usepackage{booktabs}
\usepackage{pdfcomment}
\usepackage{xcolor,colortbl}
\usepackage{array}
\usepackage{multirow}
\usepackage[T1]{fontenc}

\usepackage[utf8]{inputenc}

\usepackage{microtype}

\usepackage{inconsolata}

\newcommand\ttsmall[1]{\texttt{\small #1}}

\title{Effective Demonstration Annotation for In-Context Learning via \\ Language Model-Based Determinantal Point Process}

\author{Peng Wang$^\spadesuit$, Xiaobin Wang$^\heartsuit$, Chao Lou$^\diamondsuit$, 
\textbf{Shengyu Mao}$^\spadesuit$, \\ \textbf{Pengjun Xie}$^\heartsuit$, \textbf{Yong Jiang}$^\heartsuit$\thanks{$\quad$ Corresponding Author.}\\
$^\spadesuit$Zhejiang University,~
$^\heartsuit$Alibaba Group,~
$^\diamondsuit$ShanghaiTech University
\\
\texttt{peng2001@zju.edu.cn},~
\texttt{yongjiang.jy@alibaba-inc.com}
}

\begin{document}
\maketitle
\begin{abstract}
In-context learning (ICL) is a few-shot learning paradigm that involves learning mappings through input-output pairs and appropriately applying them to new instances.
Despite the remarkable ICL capabilities demonstrated by Large Language Models (LLMs), existing works are highly dependent on large-scale labeled support sets, not always feasible in practical scenarios.
To refine this approach, we focus primarily on an innovative selective annotation mechanism, which precedes the standard demonstration retrieval.
We introduce the \textbf{L}anguage \textbf{M}odel-based \textbf{D}eterminant \textbf{P}oint \textbf{P}rocess (LM-DPP) that simultaneously considers the uncertainty and diversity of unlabeled instances for optimal selection.
Consequently, this yields a subset for annotation that strikes a trade-off between the two factors.
We apply LM-DPP to various language models, including GPT-J, LlaMA, and GPT-3.
Experimental results on 9 NLU and 2 Generation datasets demonstrate that LM-DPP can effectively select canonical examples.
Further analysis reveals that LLMs benefit most significantly from subsets that are both low uncertainty and high diversity.
\end{abstract}

\section{Introduction}
As large pre-trained language models (LLMs) \cite{brown2020language, chowdhery2022palm, zhang2022opt, tay2023ul2, touvron2023llama, workshop2023bloom} grow in scale, 
they not only exhibit enhanced linguistic capabilities and expanded world knowledge but also demonstrate a novel ability for in-context learning. 
Specifically, LLMs have shown proficiency in learning from a limited set of input-output examples (known as demonstrations \cite{brown2020language}), and effectively applying these learned mappings to new, unseen instances.
This novel few-shot learning paradigm, which avoids parameter updates, has become a popular and efficient method for utilizing LLMs \cite{liu2021pretrain, dong2023survey, liu2021makes}.

\begin{figure}
    \centering
    \hspace*{-0.18cm}
    \includegraphics[width=0.5 \textwidth]{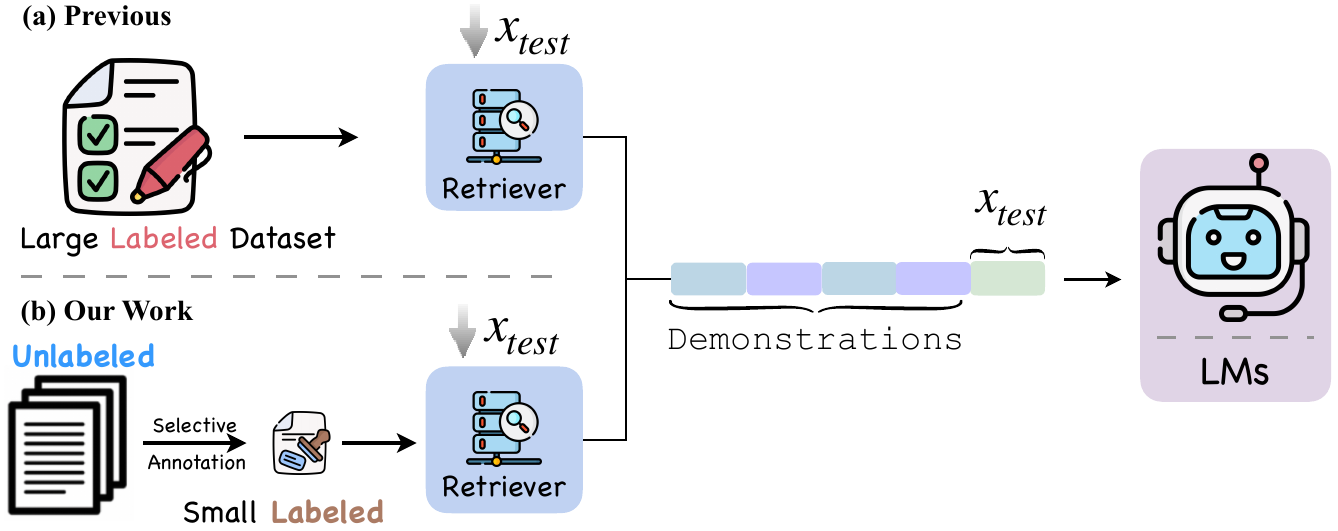}
    \caption{
    Left (\textbf{Step 1}): Without assuming access to a large amount of labeled data, we employ active data collection, selectively annotating demonstration examples. Right (\textbf{Step 2}): Prompt construction and model inference.}
    \label{fig:overview}
\end{figure}

Previous studies have investigated which instances can serve as effective prompts for ICL \cite{liu2021makes, zhang2022active, li2023finding}. 
They have demonstrated that retrieving specific similar contexts for individual test queries can significantly improve performance (\texttt{instance level}) and ground truth matters for support examples.
To assign appropriate demonstrations to all test queries, support sets necessitate diversity and broad coverage, usually achieved through large labeled data, following the principle that Monte Carlo estimation accuracy improves with larger samples.
Nonetheless, these extensive datasets are often impractical to obtain.

We investigate the selection of demonstrations from the perspective of Active Learning (AL) \cite{cohn1996active, settles2009active}.
Based on the core principle that not all data points are of equal value, AL aims to identify the most effective instances in an unlabeled data pool for annotation.
\citet{margatina2023active} elucidates that high semantic similarity, low uncertainty, and high diversity comprise an effective and efficient annotation strategy. Similarly, \citet{gonen2022demystifying} demonstrates that lower prompt perplexity is closely associated with better performance.
While \citet{su2022selective}'s Vote-k framework adopts a data-centric perspective (i.e., selecting examples that balance diversity and representativeness), it neglects the assessment of uncertainty and the inter-relationship among context examples.
In this paper, we pursue a more universally applicable yet straightforward solution, incorporating confidence signals of LLMs to select annotation instances that are maximally diverse and exhibit low uncertainty.

To address this need, we introduce a generic approach, LM-DPP, which jointly models uncertainty and diversity within the support set through a conditional Determinantal Point Process.
Specifically, we employ LLMs' perplexity to score each candidate instance in the support set, 
which serves as a measure of the LLMs' uncertainty.
Then a Gram matrix is constructed to balance the uncertainty and diversity of candidate instances and polynomial-time maximum a posteriori (MAP) inference \cite{chen2018fast} is applied to identify the most useful subset of instances to be annotated.
From the perspective of selective annotation, we consider extremely low-resource ICL scenarios as those in which the available annotated examples are limited to a few dozen instances. Our focus centers on identifying which specific set of demonstrations can most effectively harness the capabilities of LLMs within this challenging context.

We validate our method through extensive experiments on 9 NLU and 2 Generation datasets.
We also demonstrate the versatility of LM-DPP  by adapting it to the large language model GPT-3 (175B).
The experimental results illustrate that our approach can effectively balance two critical factors, uncertainty and diversity.
In summary, our contributions are as follows.
\begin{itemize}
    \item We revisit the setup of ICL from the perspective of selective annotation. We introduce a novel approach, \textbf{LM-DPP}, to select instances that balance uncertainty and diversity for annotation, aiming to reduce the human engineering workload.
    \item The experimental results indicate that the proposed method outperforms the previous best-performing selection methods by a large relative improvement and exhibits commendable generalizability across model size (\S \ref{sec:llm_scaling}) and annotation budget (\S \ref{sec:size_scaling}) scaling.
    \item Comprehensive analysis confirms that LLMs can benefit from a demonstration set that exhibits both low uncertainty and diversity (\S\ref{sec:trade-off}) and gold annotation matters for ICL performance (\S\ref{sec:ground_truth}).
\end{itemize}

\input{latex/method}

\input{latex/experiments}
\input{latex/analysis}
\input{latex/related}
\section{Conclusion and Future Work}
In this work, we focus primarily on an innovative selective annotation mechanism and introduce an efficient annotation practice, LM-DPP.
It selects both diverse and low-uncertainty examples for annotation and demonstrates promising results in various LMs.
Moreover, empirical results validate the generalizability of LM-DPP across model size and annotation budget scaling.
In the future, we plan to apply LM-DPP to more NLP tasks and explore annotation-free selection methods.

\section*{Limitations}
The proposed work still has some limitations.
\paragraph{Selection Method.}
Previous studies have elucidated that low uncertainty ensures familiarity of the LLMs with the demonstrations \cite{gonen2022demystifying}, while diversity ensures that the selected demonstrations may encompass a broad range of information, thereby enhancing the overall effectiveness of ICL \cite{margatina2023active}.
However, we still lack pilot experiments tailored to these factors to examine their impact on ICL performance thoroughly.


\paragraph{Retrieval Method.}
We have implemented prompt retrieval based on similarity (TopK).
However, it is currently unclear whether the proposed method applies to other prompt retrieval methods, such as Random Retrieval, Coverage-based Retrieval \cite{gupta2023coveragebased}, and Retrieval based on Mutual Information \cite{sorensen2022information}.
We plan to extend our work to cover more scenarios.

\paragraph{Retriever.}
Retriever is indeed one of the variables in our experiments. However, we have solely employed a retriever based on the SentenceBert architecture. Validating our experimental results on a more diverse array of retrievers constitutes future extension work.

\paragraph{Language.}
We also acknowledge that all datasets considered in this work are in English, which does not ensure that our work can be broadly generalized to other languages.

\section*{Potential Risk}
Previous works have shown Large language models contain rich biased data \cite{10.1145/3442188.3445922}.
Since we use LLMs like LlaMA, GPT-J, and GPT-3, the proposed LM-DPP approach may elicit some content with offensive language or discrimination.

\bibliography{anthology,custom}

\input{latex/appendix}

\end{document}

%% file: latex/method.tex
\section{Methodology}
\begin{figure*}[t]
    \centering
    \includegraphics[width=1.0 \textwidth]{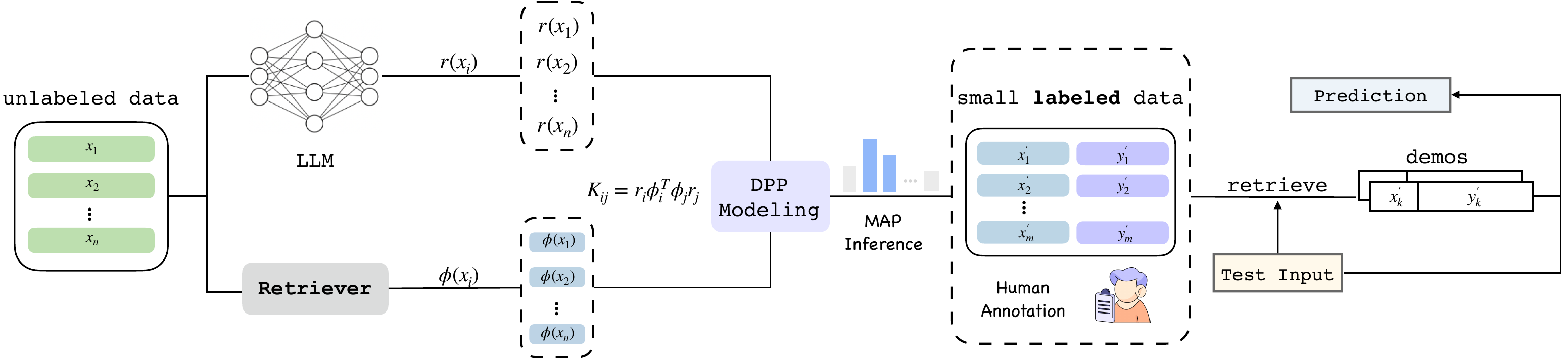}
    \caption{
    An illustration of our proposed approach. There are three steps in LM-DPP: (1) Estimate the perplexity for each unlabeled data point, with the reciprocal denoted as $r(x_i)$. (2) Employ conditional DPP to jointly model uncertainty and diversity, selecting a small set of examples for annotation before test time. (3) At test time, the context is constructed by retrieving relevant examples from the small annotated pool.
    }
    \label{fig:main_figure}
\end{figure*}
In this section, we introduce technical details of LM-DPP for selecting annotation instances exhibiting both high diversity and low uncertainty.
Formally, given a set of unlabeled samples $\mathcal{X} = \{x_i\}_{i=1}^N$, LM-DPP aims to select a subset $\mathcal{L} \subset \mathcal{X} $ for annotation, where $|\mathcal{L}| = M$ is the annotation budget, such that the Language Models (LLMs) maintains high ICL performance on the test set $\mathcal{D}_{\text{test}}$. 
As shown in Figure \ref{fig:main_figure}, given a Pre-trained Language Model (PLM) $\mathit{G}$, we first score candidate instances $x_i$ using the perplexity of the LLMs (\S\ref{sec:uncertainty}).
We then compute vector representations for the candidate instances, utilizing a conditional kernel matrix to balance diversity and low uncertainty (\S\ref{sec:dpp_modeling}). Subsequently, we perform a greedy MAP inference algorithm to filter the candidate annotation set (\S\ref{sec:inference}).

\subsection{Uncertainty}
\label{sec:uncertainty}
As off-the-shelf LLMs do not contain a classification head fine-tuned for specific tasks, calculating entropy, a common measure of uncertainty used in AL, across all possible outputs is challenging, if not unfeasible.
Alternatively, we adopt the SPELL method proposed by \cite{gonen2022demystifying}, using the perplexity of the LLMs, to score candidate examples $\widetilde{x}$.
The scoring function $r(\widetilde{x})$ is defined as:
\begin{equation}
r(\widetilde{x}) = \frac{1}{PPL(\widetilde{x})} = \exp\left(\frac{1}{t}\sum_{i=1}^{t}\log G_{\theta}(\widetilde{x}_i|\widetilde{x}_{<i})\right)
\end{equation}

Recent research also delineates that LLMs are essentially a form of lossless data compression \cite{deletang2023language}, and perplexity, serving as a proxy for the occurrence of the prompt in some form in the training data, inherently indicates the model's expectancy of the prompt.
Therefore, perplexity-based demonstration selection can, to some extent, avoid LLM sampling from low-frequency distributions. We also conduct pilot experiments (Appendix \ref{sec:high_uncertainty}) that select instances of high uncertainty, observing a substantial decrease in performance.

\subsection{DPP Modeling}
\label{sec:dpp_modeling}
We consider similarity as the primary qualitative feature of the DPP diversification process.
In this section, we present the decomposition of DPP that more directly elucidates the tension between diversity and the uncertainty measure for each candidate instance.
Since the DPP kernel, $L$ is typically written as a Gram matrix, $L = B^TB$, where the columns of $B$ represent vectors from the candidate set $\mathcal{X}$. We define $B_i$ as the product of the LLMs uncertainty term $r_i \in \mathbb{R}^+$ and the normalized diversity feature vector $\phi_i \in \mathbb{R}^D$, with $|\phi_i| = 1$. The new DPP kernel matrix can now be written as $K_{ij} = r_i \phi_i^T \phi_j r_j = r_i r_j \langle \phi_i^T \phi_j \rangle$ \cite{ye2023compositional}. $r_i$ can be regarded as the intrinsic evaluation of the LLMs for the candidate instance and $\langle \phi_i^T \phi_j \rangle$ as the measure of similarity between instances $x_i$ and $x_j$.
Therefore, we arrive at $L =\text{Diag}(r) \cdot \phi \cdot \text{Diag}(r)$, and the unnormalized log probability for the subset $S$ is $\log\text{det}(L_S) = \sum_{i \in S}\log(r_i^2) + \log\text{det}(\phi_S)$. To adjust the trade-off between uncertainty and diversity, we introduce a balancing parameter $\lambda$, thus modifying the log probability of $L_S$ to:
\begin{equation}
\label{eqn:conditional_det}
    \log \det(L_S)' = \lambda \cdot \sum\limits_{i \in S} r_i + (1 - \lambda) \cdot \log \det(L_S)
\end{equation}
This corresponds to a DPP with kernel $L' = \text{Diag}(\exp(\alpha r)) \cdot \phi \cdot \text{Diag}(\exp(\alpha r))$, where $\alpha = \lambda/(2(1 - \lambda))$.
In  Equ. (\ref{eqn:conditional_det}), the first term corresponds to the low perplexity of the selected instances, while the second term increases with the diversity of the selected instances.
Without the diversity model, we would choose examples of low uncertainty, but the DPP would tend to repeatedly select similar examples.
Without the low uncertainty model, although we could obtain a highly diverse set, we might fail to include in S those examples most favorable to the LLMs.
By combining them, we can achieve a more balanced outcome.

\subsection{Inference}
\label{sec:inference}
The solution to the MAP for DPP, which is to find the set of examples with the highest probability, is a complex process and an NP-hard problem. \cite{chen2018fast} have proposed an improved greedy algorithm that can quickly solve it approximately.
In specific, this algorithm greedily selects the demonstration from the candidate set that maximizes the marginal gain to be added to the final result subset, until the stopping condition is satisfied.
That is, each time an example $j$ is chosen to be added to the candidate set $S_{\text{map}}$, which is initialized as an empty set. The formalization is as follows:
\begin{equation}
\begin{split}
    j = \arg \max_{j\in\mathcal{X}\setminus S_{map}}&\log \text{det}(L_{S_{map} \cup \{j\}}) \\
- &\log\text{det}(L_{S_{map}})
\end{split}
\end{equation}
\input{latex/tab/basic}
By performing a Cholesky decomposition on $L_{S_{\text{map}}}$, and incrementally updating the Cholesky factor, the complexity of solving $\text{det}(L_{S_{\text{map}}})$ can be reduced from $O(K^3)$ to $O(K)$.
Therefore, the complexity of each iteration is $O(NK)$.
This implies that it is possible to return $K$ annotation examples within $O(NK^2)$ time.
Once we have selected and annotated a subset of examples$\mathcal{L}$ from the \textbf{unlabeled} support set, following recent work \cite{liu2021makes}, we retrieve examples from $\mathcal{L}$ that are semantically similar to the test query samples.
We use \verb|Sentence-BERT| \cite{reimers2019sentencebert} representations for $\mathcal{L}$ and $\mathcal{D}_{\text{test}}$ again and employ cosine similarity as the metric.
The underlying principle is that demonstrations most similar to the test example will best assist the model in answering the query.
For the order of demonstrations, we adhere to the configuration established by \citet{su2022selective}, where the order of the retrieved examples is such that $s(q_i, x) \le s(q_j, x)$ whenever $i < j$.
$s(q_i, x)$ denotes the similarity between the retrieved example $q_i$ and the test example $x$. This setup potentially leverages the recency bias inherent in LLMs \cite{zhao2021calibrate}.


%% file: latex/tab/basic.tex
\definecolor{Mycolor1}{HTML}{BAD8F2}
\definecolor{Mycolor2}{HTML}{DDEEFA}

\begin{table*}
{
\renewcommand{\arraystretch}{1.2}
\definecolor{Gray}{gray}{0.9}
\small
\resizebox{\textwidth}{!}{
\begin{tabular}{lcc|cccc|ccc|cc|c}
\toprule
\multirow{2}{*}[-0.5em]{\textbf{Model}}  & \multirow{2}{*}[-0.5em]{\textbf{Budget}} & \multirow{2}{*}[-0.5em]{\textbf{Method}} & \multicolumn{4}{c|}{\textbf{Natural Language Inference}} & \multicolumn{3}{c|}{\textbf{Classification}} & \multicolumn{2}{c|}{\textbf{Multi-Choice}} &   \multirow{2}{*}[-0.5em]{\textbf{Avg}} \\

\cmidrule{4-7} \cmidrule{8-10} \cmidrule{11-12}
&&&  RTE & MNLI & MRPC & QNLI & SST-5 & DBpedia & TREC & Hellaswag & COPA\\ 

\midrule
\multirow{10}{*}{\makecell{\textsc{\textbf{GPT-J}}\\\textsc{\textbf{6B}}}}        
& \multirow{6}{*}{$|\mathcal{L}|=16$}                   
             
&Random &48.24$_{3.1}$ &\underline{40.92$_{3.0}$} &\textbf{64.75$_{5.0}$} &51.86$_{3.5}$ &46.49$_{3.6}$ &82.72$_{7.7}$ &56.94$_{16.1}$ &67.77$_{1.5}$ &83.11$_{2.0}$ &60.31$_{6.6}$ \\
&&Kmeans &46.58$_{2.6}$ &39.84$_{1.0}$ &59.48$_{8.6}$ &51.47$_{2.1}$ &41.80$_{4.7}$ &88.77$_{0.8}$ &68.46$_{3.5}$ &66.90$_{2.2}$ &83.40$_{1.3}$ &60.74$_{3.8}$ \\
&&Vote-k &47.86$_{0.9}$ &40.04$_{2.9}$ &59.96$_{7.3}$ &51.37$_{3.9}$ &40.24$_{3.7}$ &\underline{89.26$_{3.5}$} &72.07$_{7.9}$ &68.56$_{2.9}$ &83.40$_{1.6}$ &61.42$_{4.4}$ \\
&&Fast Vote-k &48.34$_{0.7}$ &39.26$_{3.9}$ &58.89$_{5.0}$ &50.39$_{1.7}$ &\textbf{50.80$_{5.8}$} &\textbf{89.65$_{3.4}$} &75.10$_{5.5}$ &67.38$_{3.8}$ &83.10$_{0.8}$ &62.54$_{3.8}$ \\
&&\colorbox{Mycolor2}{LM-DPP (ours)} &\textbf{49.81$_{1.5}$} &\underline{40.92$_{1.7}$} &\underline{64.36$_{1.4}$} &\textbf{52.96$_{2.0}$} &\underline{47.66$_{5.0}$} &89.06$_{3.0}$ &\textbf{75.20$_{2.6}$} &\textbf{69.44$_{2.6}$} &\textbf{83.60$_{2.1}$} &\textbf{63.67$_{2.6}$} \\

\cmidrule{2-13} 
& \multirow{6}{*}{$|\mathcal{L}|=100$}     
&Random &47.64$_{2.2}$ &39.41$_{2.8}$ &63.59$_{3.1}$ &51.11$_{3.5}$ &47.43$_{0.9}$ &90.30$_{1.5}$ &76.36$_{1.3}$ &67.88$_{0.8}$ &\textbf{84.03$_{1.7}$} &63.08$_{2.2}$ \\
&&Kmeans &48.22$_{0.5}$ &41.74$_{3.8}$ &64.40$_{5.0}$ &51.52$_{3.1}$ &46.18$_{1.6}$ &90.55$_{1.7}$ &77.09$_{5.6}$ &67.63$_{0.5}$ &83.30$_{1.8}$ &63.40$_{3.1}$ \\
&&Vote-k &49.12$_{1.3}$ &40.26$_{2.9}$ &61.24$_{4.1}$ &50.62$_{3.1}$ &47.85$_{1.2}$ &86.92$_{2.0}$ &\textbf{82.18$_{2.5}$} &67.79$_{1.8}$ &82.12$_{2.8}$ &63.12$_{2.6}$ \\
&&Fast Vote-k &51.93$_{4.1}$ &39.53$_{4.2}$ &65.73$_{1.2}$ &50.41$_{2.6}$ &49.39$_{0.9}$ &\textbf{91.60$_{2.1}$} &81.45$_{5.4}$ &68.23$_{1.0}$ &\underline{83.84$_{3.9}$} &64.68$_{3.2}$ \\
&&\colorbox{Mycolor2}{LM-DPP (ours)} &\textbf{54.44$_{2.6}$} &\textbf{42.31$_{2.4}$} &\textbf{67.10$_{1.3}$} &\textbf{53.26$_{1.5}$} &\textbf{49.62$_{1.0}$} &\underline{91.03$_{2.2}$} &\underline{82.01$_{3.2}$} &\textbf{68.92$_{1.5}$} &83.80$_{1.7}$ &\textbf{65.83$_{2.0}$} \\

\midrule
\multirow{10}{*}{\makecell{\textsc{\textbf{LlaMA-2}}\\\textsc{\textbf{7B}}}} 
& \multirow{6}{*}{$|\mathcal{L}|=16$}                    

&Random &54.70$_{1.4}$ &\underline{38.81$_{1.4}$} &60.42$_{1.9}$ &53.03$_{2.1}$ &54.10$_{4.1}$ &86.82$_{6.0}$ &67.48$_{14.4}$ &\underline{77.25$_{2.1}$} &88.58$_{2.5}$ &64.57$_{5.6}$ \\
&&Kmeans &54.88$_{1.3}$ &36.62$_{4.9}$ &60.94$_{8.0}$ &52.54$_{1.8}$ &53.32$_{2.7}$ &90.04$_{1.8}$ &\underline{76.95$_{8.4}$} &\underline{77.25$_{2.1}$} &\textbf{89.06$_{1.4}$} &65.73$_{4.5}$ \\
&&Vote-k &52.83$_{0.5}$ &\textbf{41.21$_{4.8}$} &62.89$_{1.3}$ &\textbf{55.57$_{0.4}$} &53.42$_{2.6}$ &87.79$_{1.6}$ &\textbf{79.10$_{2.5}$} &77.24$_{2.4}$ &87.70$_{1.3}$ &66.42$_{2.3}$ \\
&&Fast Vote-k &52.25$_{1.2}$ &38.28$_{4.0}$ &59.67$_{4.4}$ &53.13$_{1.7}$ &53.32$_{4.3}$ &88.28$_{1.8}$ &75.46$_{4.7}$ &77.15$_{2.9}$ &88.48$_{1.9}$ &65.11$_{3.3}$ \\
&&\colorbox{Mycolor2}{LM-DPP (ours)} &\textbf{58.99$_{3.5}$} &38.28$_{5.6}$ &\textbf{63.09$_{4.5}$} &\underline{53.81$_{2.6}$} &\textbf{55.37$_{3.3}$} &\textbf{93.65$_{1.5}$} &76.28$_{4.5}$ &\underline{77.25$_{1.2}$} &\underline{88.67$_{1.1}$} &\textbf{67.26$_{3.5}$} \\

\cmidrule{2-13} 
& \multirow{6}{*}{$|\mathcal{L}|=100$}     

&Random &58.01$_{1.2}$ &39.85$_{5.1}$ &60.48$_{4.0}$ &51.66$_{1.9}$ &54.50$_{1.6}$ &92.87$_{1.2}$ &83.69$_{2.6}$ &76.76$_{3.1}$ &87.91$_{1.2}$ &67.30$_{2.8}$ \\
&&Kmeans &56.54$_{1.3}$ &\textbf{42.29$_{2.9}$} &64.85$_{2.2}$ &53.32$_{2.1}$ &54.78$_{1.9}$ &93.75$_{2.0}$ &\underline{84.96$_{2.9}$} &78.03$_{2.3}$ &87.70$_{1.5}$ &68.47$_{2.2}$ \\
&&Vote-k &58.40$_{0.7}$ &\underline{42.19$_{3.2}$} &65.33$_{4.0}$ &53.71$_{1.4}$ &57.13$_{2.3}$ &90.82$_{1.5}$ &84.38$_{2.7}$ &78.42$_{3.3}$ &86.14$_{1.6}$ &68.50$_{2.5}$ \\
&&Fast Vote-k &61.72$_{0.3}$ &39.55$_{1.5}$ &63.18$_{1.4}$ &51.95$_{1.0}$ &56.15$_{2.1}$ &93.46$_{0.7}$ &\textbf{85.74$_{1.9}$} &77.83$_{3.0}$ &88.18$_{1.5}$ &68.64$_{1.7}$ \\
&&\colorbox{Mycolor2}{LM-DPP (ours)} &\textbf{58.99$_{2.7}$} &41.31$_{5.3}$ &\textbf{66.80$_{2.3}$} &\textbf{56.15$_{0.9}$} &\textbf{57.62$_{3.0}$} &\textbf{94.82$_{0.4}$} &83.50$_{2.2}$ &\textbf{78.91$_{2.1}$} &\textbf{89.36$_{1.8}$} &\textbf{69.72$_{2.6}$} \\
\bottomrule
\end{tabular}
}
}
\caption{Results with GPT-J and LlaMA-2-7B on NLU task. We compare various selective annotation methods with $\{100, 16\}$ annotated examples. Bold numbers indicate the highest accuracy among all methods, while those underlined indicate the second-best. The subscript denotes the standard deviation.}
\label{tab:main_table}
\end{table*}

%% file: latex/experiments.tex
\section{Experiments}
\subsection{Experimental Settings}
\paragraph{Datasets}
We conduct experiments on 9 NLU and 2 Generation tasks involving different task formulations, including \textbf{Sentiment Classification}: SST-5 \cite{socher-etal-2013-recursive}; \textbf{Natural Language Inference}: RTE \cite{bentivogli2009fifth}, MNLI \cite{williams2017broad}, MRPC \cite{dolan2004unsupervised}, QNLI \cite{wang2018glue}; \textbf{Topic Classification}: TREC \cite{hovy2001toward}, DBpedia \cite{lehmann2015dbpedia}; \textbf{Multiple-choice Question Answering}: Hellaswag \cite{zellers2019hellaswag}, COPA \cite{roemmele2011choice}; \textbf{Abstractive Summarization}: XSUM \cite{narayan2018dont} and \textbf{Open Domain QA}: NQ \cite{kwiatkowski-etal-2019-natural}.
In the main experiment, the budget of annotation is set as ($\{16, 100\}$).
For datasets with publicly available test data, we use the test data for evaluation. For others, we follow previous work \cite{lan2019albert, su2022selective} and use the dev set for evaluation.

\paragraph{Baselines}
We compare LM-DPP with four strong selective annotation methods. And in our study, we primarily utilize \textbf{GPT-J-6B} \cite{gpt-j} and \textbf{LlaMA-2-7B} \cite{touvron2023llama} as scoring and inference language models, More details about baselines and implementation can be found in Appendix \ref{sec:baselines}, \ref{sec:implementation_detail} respectively.

\paragraph{Metrics}
\input{latex/tab/generation_task}
We compare the predicted answers with the true outcomes and report the accuracy (Acc.) for all NLU tasks and exact matching scores \cite{rajpurkar-etal-2016-squad} for NQ. 
For summarization tasks, we assess factual consistency using FactCC \cite{kryscinski-etal-2020-evaluating} \footnote{\url{https://huggingface.co/manueldeprada/FactCC}}, a BERT-based \cite{devlin-etal-2019-bert} metric for evaluating output faithfulness. Simultaneously, for quality assessment, we report the ROUGE-L F1 score \cite{lin-2004-rouge} to evaluate the summary against the reference.

\subsection{Main Results}
\label{sec:main_results}
\paragraph{NLU Task}
From Table \ref{tab:main_table}, we can observe that LM-DPP consistently improves the \textbf{on-average} accuracy across a variety of NLU tasks under different annotation budgets ($|\mathcal{L}|=16$, $|\mathcal{L}|=100$).
Specifically, with a larger budget, LM-DPP achieves an average absolute gain of 1.15\% on GPT-J and 1.08\% on LlaMA, compared to the best-performing baseline.
This demonstrates that balancing uncertainty and diversity ensures that the chosen demonstrations are more likely to contain complementary information that enhances performance.
On GPT-J, LM-DPP exhibits the lowest average standard deviation (2.6, 2.0), and on LlaMA-2, it shows greater stability than the Random baseline, albeit marginally lower than Vote-k. 
This indicates that LM-DPP can maintain a relatively stable performance across different experimental setups, substantially increasing the reliability and robustness of contextual learning.
Furthermore, we observe that as the annotation budget increases, performance fluctuations decrease across different selection methods.

\paragraph{Generation Task}
Experiments on LlaMA-2 (as shown in Table \ref{tab:generation_task}) reveal that LM-DPP achieves notable improvement on the NQ task across various annotation budgets, especially at $\mathcal{L}=16$, where it surpasses the best baseline by 1.04\%.
In the XSUM task, applying LM-DPP consistently enhances Rouge scores, particularly achieving a 2.18\% increase at $\mathcal{L}=100$. 
This underscores the efficacy of the proposed method in improving the generality and reference similarity of generated text.
However, this improvement comes \textbf{at the cost of some degree of factual consistency} with the reference, potentially due to the pursuit of diversity reducing the focus on task-specific relevance (see Appendix \ref{sec:fact_consistency_analysis} for a more detailed analysis). 
Overall, LM-DPP boosts the model's generalization and accuracy and highlights the potential for performance optimization with increased annotation budgets. 
Despite some variability in factual consistency, these insights pave the way for future research on efficiently allocating annotation resources in NLG tasks \cite{Dong_2022}.
\begin{figure}[htbp]
    \centering
    \hspace*{-0.20cm}
    \includegraphics[width=0.5 \textwidth]{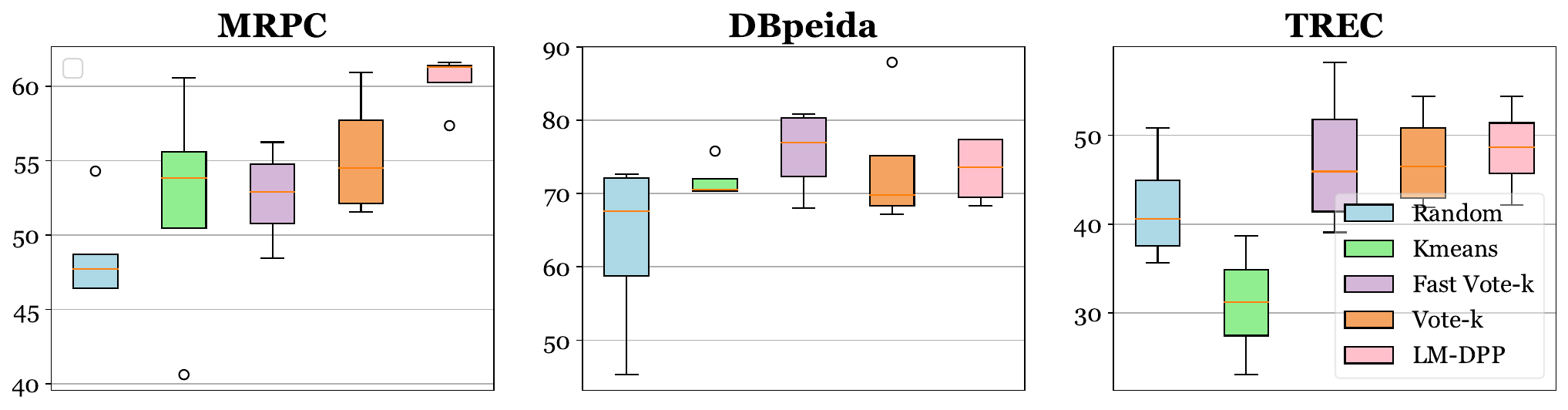}
    \caption{LlaMA-2-7B Results with $\mathcal{L} = 4$.}
    \vspace{-0.25cm}
    \label{fig:4_shot}
\end{figure}
\paragraph{Smaller In-Context Examples}
We investigate the impact of the number of examples and labels on ICL performance. As shown in Figure \ref{fig:4_shot}, LM-DPP surpasses the other baselines in terms of accuracy and stability on MRPC and TREC but is slightly inferior to Vote-k on DBpedia.
Further analysis suggests that a well-balanced demonstration set does not always result in improved performance or reduced variance (see Appendix \ref{sec:label_coverage} for more details). In TREC, performance increases with more labels, whereas in MRPC, demonstrations with a single label (all being \texttt{equivalent}) lead to better performance than a balanced demonstration set, with less variance.



%% file: latex/tab/generation_task.tex

\begin{table}[t]
\centering
\resizebox{1.0 \columnwidth}{!}{
\begin{tabular}{ll|ccccc}
\toprule
\multicolumn{2}{c|}{\textbf{Methods}}  & Random & Kmeans &Vote-k &Fast Vote-k &LM-DPP \\

\midrule
\multicolumn{7}{c}{\textsc{$\mathcal{L}$ = 16}} \\
\midrule

\textbf{NQ} &\textit{ACC.}	&21.74$_{4.39}$ &22.78$_{3.63}$ &\underline{22.79}$_{3.37}$ &22.01$_{3.75}$	&\textbf{23.83}$_{3.10}$ \\
\midrule

\multirow{2}{*}{\textbf{XSUM}} &\textit{R-L} &24.57$_{0.03}$ &23.65$_{0.29}$ &\underline{24.88}$_{1.03}$ &24.74$_{1.20}$ &\textbf{26.34}$_{1.07}$ \\

&\textit{FactCC} &\underline{35.07}$_{4.26}$ &\textbf{36.72}$_{2.41}$ &32.49$_{1.44}$ &34.68$_{2.86}$ &33.53$_{3.70}$ \\

\midrule
\multicolumn{7}{c}{\textsc{$\mathcal{L}$ = 100}} \\
\midrule

\textbf{NQ} &\textit{ACC.} &23.57$_{3.54}$ &22.92$_{3.13}$ &\underline{24.48}$_{4.01}$	&23.70$_{3.51}$ &\textbf{24.61}$_{3.74}$ \\
\midrule

\multirow{2}{*}{\textbf{XSUM}} &\textit{R-L} &\underline{25.11}$_{0.41}$	&24.47$_{0.46}$ &24.66$_{0.84}$ &24.63$_{1.37}$ &\textbf{27.29}$_{0.55}$ \\

&\textit{FactCC} &35.64$_{5.86}$ &34.86$_{2.97}$ &\underline{36.12}$_{2.40}$ &\textbf{36.53}$_{3.84}$ &35.16$_{2.01}$ \\

\bottomrule
\end{tabular}
}
\caption{Results with LlaMA-2-7B on Generation Task.}
\vspace{-0.25cm}
\label{tab:generation_task}
\end{table}

%% file: latex/analysis.tex
\section{Analysis}
\subsection{Impacts of the Trade-off Between Uncertainty and Diversity}
\label{sec:trade-off}

We analyze to investigate how the trade-off between diversity and uncertainty impacts the performance of downstream tasks.
With an annotation budget of 100, we test the performance under different $(\lambda)$ values utilizing GPT-J as the inference model.
As evident from Table \ref{tab:trade-off}, a complete inclination towards uncertainty $(\lambda=1.0)$ generally yields poorer outcomes across all tasks, likely due to selective annotation excessively concentrating on a small portion of data, thereby diminishing ICL's generalization capacity.
\input{latex/tab/trade-off}
\begin{figure*}[t]
    \centering
    \includegraphics[width=1.0\textwidth]{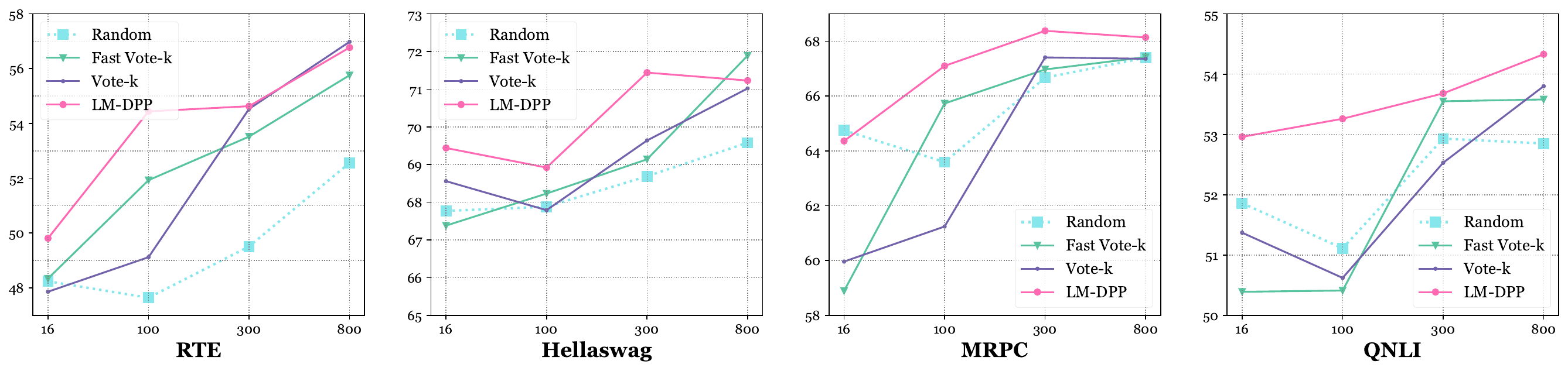}
    \caption{Comparisons of various selection methods with (\{16, 100, 300, 800\}) annotated examples on four representative tasks: RTE, MRPC paraphrase detection, QNLI, and Hellaswag commonsense answering for GPT-J.
    }
    \label{fig:budget_scaling}
\end{figure*}
Optimal effects are often observed at $(\lambda)$ values of \textbf{0.5} or \textbf{0.6} (which approximate a balance between the two factors), suggesting that moderate uncertainty coupled with a degree of diversity is beneficial for the model's downstream task performance. 
Moreover, different tasks demonstrate varied sensitivities to the $(\lambda)$ value. 
For instance, QNLI shows minor performance shifts ($\pm1.95\%$), whereas DBpedia exhibits significant performance variations at certain $(\lambda)$ values (exceeding $\pm10.00\%$), indicating that the optimal selection of $(\lambda)$ may relate to the tasks' characteristics and difficulty levels.
Despite such variability, we find that introducing this trade-off factor consistently surpasses the vanilla DPP and Perplexity baselines, which consider only diversity or uncertainty, thereby validating the effectiveness of LM-DPP.

\subsection{Transferability across Different LMs}
\label{sec:llm_scaling}

\begin{figure}
    \hspace*{-0.32cm}
    \includegraphics[width=0.5 \textwidth]{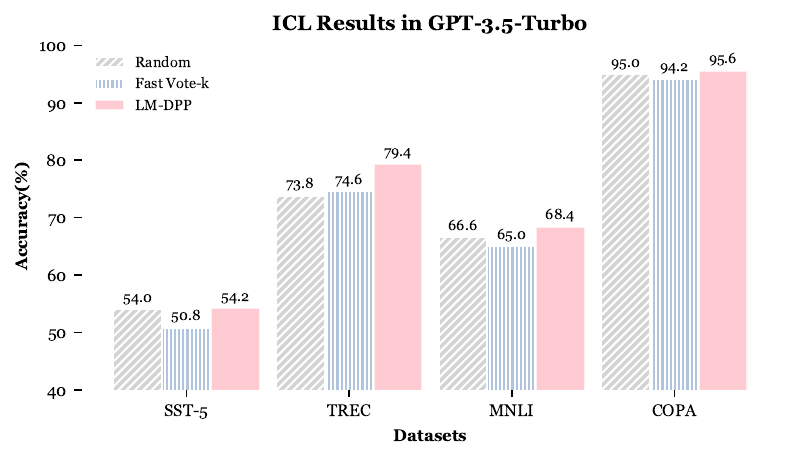}
    \caption{Results of GPT-3-Turbo (175B) with 100 annotated examples. LM-DPP consistently improves in-context learning on various datasets.}
    \label{fig:transfer_gpt3}
\end{figure}

\paragraph{Small model for scoring}
Scoring every sample from the extensive unlabeled pool using a more resource-intensive LLM could be computationally demanding, particularly when the size of the unlabeled sample pool is substantial. Therefore, we attempt to use GPT2 \cite{radford2019language} (117M, which possesses basic language modeling capabilities) as a surrogate for the source language model GPT-J, while maintaining GPT-J for the inference model.
Across 9 NLU tasks (annotation size=100), the average accuracy was \textbf{64.76} (details in Appendix \ref{sec:small_model_score}).
This indicates that LM-DPP exhibits strong transferability across different inference LMs, which means that the selected demonstrations can be reused.

\paragraph{Transfer to LLMs}

To gain some intuition on the effect of model size, we endeavor to transfer the proposed method to LLMs that are aligned with human expectations (\ttsmall{gpt-3.5-turbo-instruct}) \cite{ouyang2022training}.

In specific, we take the \emph{logprobs} returned by the official API as a reference for measuring uncertainty, from which we calculate $r(x_i)$ and perform standard LM-DPP.
As depicted in Figure \ref{fig:transfer_gpt3}, we report the experimental results of GPT-3.5-Turbo (175B) with LM-DPP on several datasets and compare them with the Random and Fast Vote-k baseline.
In comparison to random selection, our results indicate that LM-DPP can significantly enhance the performance of GPT-3.5, as evidenced by the 5.6\% improvement in TREC accuracy, 1.8\% in MNLI, 0.2\% in SST-5, and 0.6\% in COPA.
The proposed LM-DPP approach surpasses Fast Vote-k by an average of 3.25\%, indicating that considering representativeness alone is not sufficient to extract a high-quality demonstration subset.

\subsection{Varying budget of annotated examples}
\label{sec:size_scaling}
We further investigate how the size of the annotation set affects the performance of in-context learning. 
Under annotation sizes of (\{16, 100, 300, 800\}), we compare LM-DPP with Random selection, Fast Vote-k, and Vote-k, and report the results in Figure \ref{fig:budget_scaling}.
It is observable that with increasing annotation budgets, most selective methods generally show a consistent overall improvement trend.
This is in line with the expectation that more labeled data is more likely to retrieve relevant examples to assist LLMs in accurately answering, thereby improving the performance of in-context learning.
The proposed approach, LM-DPP, outperforms other methods at an annotation size of 16 on RTE, Hellaswag, and QNLI, suggesting that even with extremely low annotation budgets, LM-DPP can ensure the effectiveness and diversity of context. 
Additionally, with a sufficient annotation budget ($\mathcal{L} = 800$), LM-DPP exhibits commendable performance, achieving the best results on two datasets, MRPC and QNLI.
In contrast, the performance decline of Vote-k on QNLI may be attributed to the annotation of noisy data (high perplexity), with some case analyses provided in the appendix \ref{sec:appendix_perplexity}.
This reaffirms the necessity of balancing uncertainty and diversity.

\subsection{Time Efficiency}
\begin{figure}[t]
    \centering
    \hspace*{-0.35cm}
    \includegraphics[width=0.53 \textwidth]{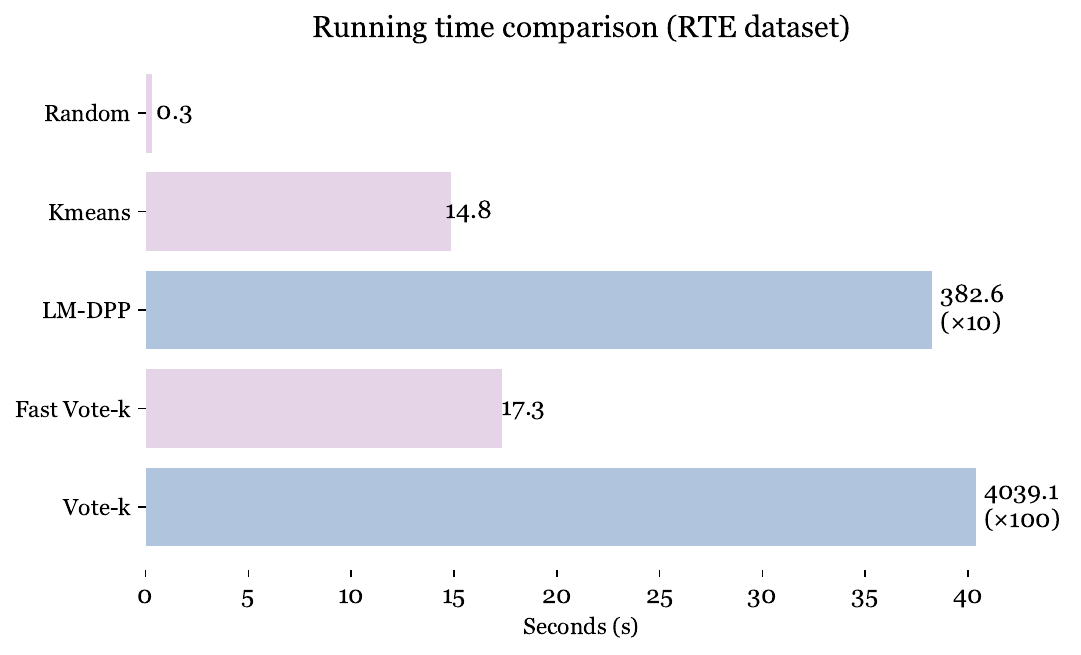}
    \caption{The time consumed to select 300 demonstrations from the RTE dataset (comprising 2491 instances).}
    \label{fig:time_cost}
\end{figure}

We explore the execution efficiency of both the baseline methods and LM-DPP.
As illustrated in Figure \ref{fig:time_cost}, the LM-Free approach significantly reduces the time required to select demonstrations compared to methods that require scoring by LM.
Selecting 300 samples takes 4039.1s with Vote-k, 382.6s with LM-DPP, and only 0.3s with random selection.
Since LM-DPP only requires a single forward pass per sample, we can optimize time efficiency in two ways: (1) preemptively compute perplexity for data samples in practical scenarios and devise methods to reset or update cached demonstration samples periodically. (2) using smaller-parameter scoring models (see \S \ref{sec:llm_scaling}) can achieve more than tenfold acceleration (24.4s).

\section{Discussion}
\subsection{Case Study}
\begin{figure}[t]
    \centering
    \hspace*{-0.30cm}
    \includegraphics[width=\columnwidth]{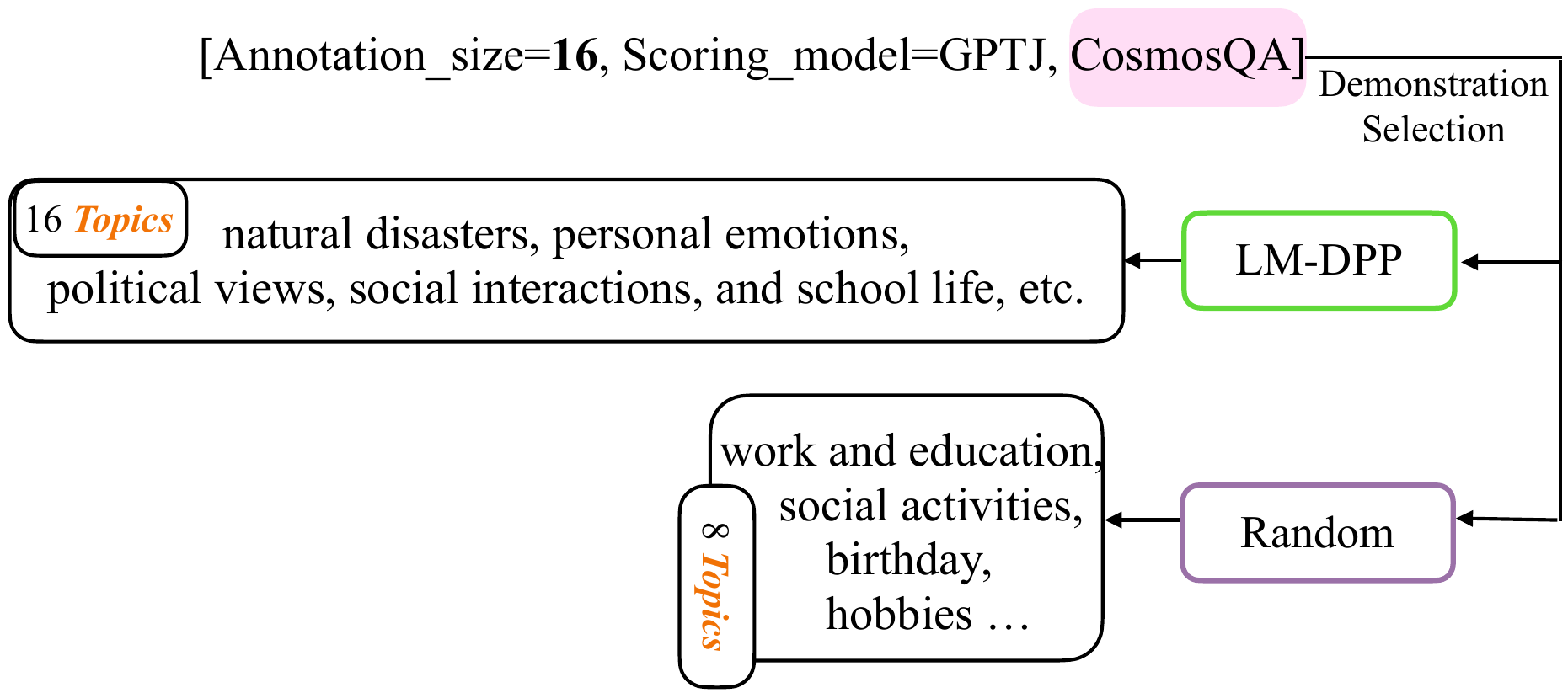}
    \caption{Case Study of selected demonstrations under the condition of annotation\_size=16.}
    \label{fig:case_study}
\end{figure}

We compare demonstrations selected via LM-DPP against Random in CosmosQA dataset \cite{huang-etal-2019-cosmos}. 
It reveals that demonstrations selected by the LM-DPP exhibit greater diversity in content, covering 16 distinct topics such as natural disasters, personal emotions, political views, social interactions, and school life, compared to only 8 topics covered by random selection (Figure \ref{fig:case_study}). 
The selected demonstrations not only span a broad range of subjects but also offer a variety in style, including personal narratives, descriptive events, emotional expressions, and dialogues. 
This diversity enhances the model's ability to interpret and respond to questions.

\subsection{Does annotation benefit from gold labels?}
\label{sec:ground_truth}
\citet{min2022rethinking} observed that random substitution of labels in demonstrations minimally impacts the performance across a suite of tasks, while \citet{yoo2022groundtruth} highlighted that the integrity of input label mapping is a crucial factor.
In this section, we explore whether Gold Labels (i.e., providing correct labels) are essential for achieving high performance in ICL.

Specifically, we divide the selective annotation process into several steps.
Step 1: Annotate 50 instances to construct an in-domain dev set $\mathcal{D}^{i}$ (containing gold labels).
Step 2: For the unannotated instances, we pair each input $x_i$ with every possible label $y \in \mathcal{C}$ ($\mathcal{C}$ is the label set) to construct a train set $\mathcal{D}'$ carrying pseudo-labels.
Step 3: Given the prompts $\mathcal{Z} \in \mathcal{D}'$, the ICL accuracy on the in-domain dev set $\mathcal{D}^{i}$ is denoted as $\text{Acc}(\mathcal{Z})$. We select the Top-50 $\mathcal{Z}$, represented as $\mathcal{D}^u$.
Therefore, the final annotation set ($|\mathcal{L}| = 100$) comprises two parts: $\mathcal{D}^{i}$ with gold labels, and $\mathcal{D}^u$ selected post-hoc.
This process is referred to as \textbf{UN-LM-DPP}, followed by conducting standard ICL experiments.

\input{latex/tab/in-dev}
As shown in Table \ref{tab:in-dev}, we observe that UN-LM-DPP, compared to LM-DPP with gold annotations, exhibits a certain performance decline in most tasks but still surpasses Random selection in some datasets.
The performance fluctuation varies significantly across different tasks, depending on the specific characteristics of the datasets, as evidenced by a decrease of -6.74\% in TREC, yet only -2.88\% in MNLI.

\input{latex/tab/correct-label-rate}
This suggests that, to a certain extent, ICL generally benefits from gold demonstrations.
In addition, we report the proportion of gold demonstrations within the constructed $\mathcal{D}^u$ during Step 2, with the results presented in Table \ref{tab:correct-label}.
In QNLI, there is a 52.30\% gold label ratio, and surprisingly, we observe a slight performance improvement compared to LM-DPP.
It is evident that within similar tasks, a higher ratio of gold-standard examples correlates with a smaller decline in ICL performance.
However, this is not a generalized finding across the board, and we consider annotation-free ICL as a direction for future work.

%% file: latex/tab/trade-off.tex
\begin{table}[t]
\resizebox{\linewidth}{!}{
\begin{tabular}{lccccc}
\toprule
$\lambda$   & MRPC & QNLI & TREC & DBpedia & Hellaswag  \\ 

\cmidrule{2-6}

0.0	&62.57	&51.43	&79.40	&90.67	&67.16 \\
0.2	&66.42	&52.64	&78.82	&89.47	&66.73 \\
0.4	&65.34	&53.21	&77.69	&90.22	&65.05 \\
0.5	&\underline{66.89}	&\textbf{53.38}	&81.43	&\textbf{91.52}	&\underline{68.89} \\
0.6	&\textbf{67.10}	&\underline{53.26}	&\textbf{82.01}	&\underline{91.03}	&\textbf{68.92} \\
0.8	&66.39	&52.18	&81.24	&90.77	&67.42 \\
0.9	&66.51	&52.97	&79.36	&84.25	&66.27 \\
1.0	&66.14	&51.45	&\underline{81.57}	&79.49	&59.73 \\
 \bottomrule
\end{tabular}
}
\caption{ The GPT-J performance of different trade-off factors $\lambda$. $(\lambda = \{0.0, 1.0\})$ correspond respectively to the vanilla DPP and the Perplexity baseline (\S \ref{sec:baselines}).
}
\label{tab:trade-off}
\end{table}

%% file: latex/tab/in-dev.tex

\begin{table}[t]
\def\arraystretch{1.4}
\resizebox{\linewidth}{!}{
\begin{tabular}{l||cccccc}
\toprule[1.2pt]          
&    Hellaswag &  COPA  &  DBpedia  &    TREC &    QNLI   &   MNLI  \\
\midrule                                                                                   

Random$\dagger$        & 67.88 & 84.03 & 90.30 & 76.36 & 51.11 & 39.41 \\ 
LM-DPP$\dagger$      & 68.92 & 83.80 & 91.03 & 82.01 & 53.26 & 42.31\\
\midrule
UN-LM-DPP    & $68.48_{\hspace{0.05cm}\textbf{-0.64}}$ & $83.20_{\hspace{0.05cm}\textbf{-0.72}}$ & $90.74_{\hspace{0.05cm}\textbf{-0.32}}$ & $76.48_{\hspace{0.05cm}\textbf{-6.74}}$ & $53.37_{\hspace{0.05cm}\textbf{+0.21}}$ & $41.09_{\hspace{0.05cm}\textbf{-2.88}}$ \\ 

\bottomrule[1.2pt]
\end{tabular}
\vspace{-0.8cm}
}
\caption{The GPT-J performance on various datasets. $\dagger$Resulting numbers are taken from Table \ref{tab:main_table}. The annotation budget is 100. In UN-LM-DPP, the annotation set consists of two parts: $\mathcal{D}^{i}$ and $\mathcal{D}^{u}$, with standard ICL being implemented.
}
\label{tab:in-dev}%
\end{table}

%% file: latex/tab/correct-label-rate.tex
\begin{table}[htbp]
\centering
\setlength{\tabcolsep}{4pt}
\def\arraystretch{1.25}
\resizebox{\linewidth}{!}{
\begin{tabular}{c|cccccc}
\toprule[1.2pt]
\textbf{Dataset}    & Hellaswag    & COPA    & DBpedia    & TREC    & QNLI & MNLI    \\ \hline
\textbf{Gold-Labeled} & 47.63\% & 38.86\% & 25.11\% & 11.52\% & 52.30\% &37.43\% \\
\bottomrule[1.2pt]
\end{tabular}%
\vspace{-0.3cm}
}
\caption{The proportion of golden-labeled examples identified within an unlabeled setting in UN-LM-DPP.}
\label{tab:correct-label}
\end{table}

%% file: latex/related.tex
\section{Related Work and Background}

\paragraph{Determinantal Point Process}
The Determinantal Point Process (DPP) is an elegant probabilistic model that captures negative correlations and allows for efficient algorithms in sampling, marginalization, and conditioning \cite{Kulesza_2012}.
Formally, a point process $\mathcal{P}$ is a probability measure on the power set of $\mathcal{V}$, that is, the set of all discrete items $2^\mathcal{V}$. If $\mathit{Y}$ is a random subset drawn according to $\mathcal{P}$, then for every $S \subseteq \mathit{Y}$:

\begin{equation}
P(S \subseteq \mathit{Y}) = \text{det}(L_S)
\end{equation}
for some kernel matrix $L \in \mathbb{R}^{n \times n}$ that is symmetric, real and positive semidefinite.
$L_S$ denotes the submatrix of $L$ obtained by restricting to the rows and columns indexed by $S$.
The operator $\text{det}(\cdot)$ represents the determinant of a matrix.
Typically, the DPP kernel $L$ can be written as a Gram matrix, $L_{ij} = K(a_i, a_j)$, where $K(\cdot,\cdot)$ is the kernel associated with the determinantal point process, often expressed as $\phi(a_i)^T \phi(a_j)$, $\phi$ is the feature map of a reproducing kernel \cite{ye2023compositional}.

Under distribution $\mathcal{P}$, our objective is maximum a posteriori (MAP) inference, which is to find the subset of items with the highest probability, corresponding to the most diverse subset of items.
\begin{equation}
S_{map} = \arg \max_{S \in \mathit{Y}}\text{det}(L_S)
\end{equation}
Although finding the mode of a DPP is NP-hard, pioneering works \cite{Kulesza_2012, lee2009nonmonotone, chen2018fast, NIPS2012_6c8dba7d} have largely relied on greedy algorithms or sampling methods, and have succeeded in performing greedy MAP inference within polynomial time.

\paragraph{In-context Learning}
The capacity for in-context learning has been observed in large-scale Pre-trained Language Models (PLMs) such as GPT-3, representing a few-shot learning paradigm that does not require any parameter updates.
It involves pre-pending a small number of demonstrations as prompts before the test input, allowing LLMs to discern patterns and ``learn'' to predict.

Formally, let $\hat{x}$ be the test query to be addressed, and $s(\cdot,\cdot)$ be the cosine similarity.
Standard ICL prompts the language model $\mathit{G}$ with a set of example input-output pairs $\{(x_1, y_1) \dots (x_m, y_m)\}$ and predicts the answer $\hat{y}$ for the query.
Typically, the pairs $(x_i, y_i)$ are retrieved from a train set $\mathcal{D}$ within the same domain through similarity.
\begin{align}
\hat{y} = \arg \max_{y}G_{\theta} (y \mid \hat{x}, \mathcal{C}),\\
\mathcal{C} = \mathop{\text{TopK}}\limits_{(x_i, y_i) \in \mathcal{D}} (s(\hat{x}, x_i)).\nonumber
\end{align}
Recent works have aimed to enhance ICL by selecting valuable demonstrations \cite{liu2021makes, rubin2022learning}, optimizing the order of demonstrations \cite{lu2022fantastically}, etc.
\citet{su2022selective} utilize selective annotation to significantly reduce annotation costs while ensuring high ICL performance. 
\citet{yang-etal-2023-representative} explore the corpus-level in-context learning via DPP and mention the need to use gold labels to score candidate samples. CEIL \cite{ye2023compositional} train the demonstration retriever with a learnable conditional DPP. However, these existing works are highly dependent on large annotated support sets.

%% file: latex/appendix.tex
\appendix

\section{Appendix}
\label{sec:appendix}

\subsection{Details with perplexity estimation}
\label{sec:appendix_perplexity}
\input{latex/tab/appendix1}

We report the perplexity of annotated instances when ($|\mathcal{L}| = \{16, 100\}$) (as shown in Table \ref{tab:appendix1}).
It's observed that as the annotation cost increases to 100, there is a corresponding significant rise in perplexity.
For instance, in COPA, the $\text{Perplexity}_{\text{avg}}$ increases by 4.01, and $\text{Perplexity}_{\text{max}}$ rises by 125.70. A similar phenomenon is also observed in DBpedia.
This indicates to some extent that introducing demonstrations with high perplexity can lead to a decrease in ICL performance.

\subsection{Implementation Details}
\label{sec:implementation_detail}

The inference method we employed is direct (a regular inference used in \cite{brown2020language}), which involves presenting demonstrations and candidate answers to the LLMs to select the candidate with the highest likelihood.
For each test dataset, a specific prompt template (Table \ref{tab:prompt_template}) is used for scoring and inference.
For each test instance, we include as many retrieved samples as possible in the preceding prompt, up until the maximum token length was reached (e.g., 2048 for GPTJ, 4096 for LlaMA-2-7B).
Sentence-BERT \cite{reimers2019sentencebert} is used as the demonstration retriever.
\input{latex/tab/dataset}
Following \cite{rubin2022learning}, we adopt the \ttsmall{paraphrase-mpnet-base-v2} to encode the test input $x_{\text{test}}$ and the inputs of the train set.
All experiments are conducted on a single Tesla V100 GPU with 32GB of memory. 
Empirically, obtaining embeddings for unlabeled examples using \texttt{Sentence BERT} as described in Section \ref{sec:uncertainty} varies between 0.2 to 2 hours, contingent upon the dataset size.
In Section \ref{sec:dpp_modeling}, our approach requires approximately 6 seconds to generate the annotation set on a single CPU.
Notably, ICL obviates the need for model training.

We also acknowledge that acquiring unlabelled samples in practice is a process marked by significant variance\cite{su2022selective}. To simulate this realistic scenario, \textbf{we randomly sample 3K instances} from the training set multiple times to serve as the pool of samples awaiting annotation.
In all the experimental setups described in this paper, we utilize four distinct seeds \textbf{(0, 1, 42, 123)}, and the values presented in the tables (figures) reflect the average across four runs. Additionally, we provide the corresponding standard deviations for these values.

\subsection{Baselines}
\label{sec:baselines}
\begin{figure*}[t]
    \centering
    \includegraphics[width=1.0 \textwidth]{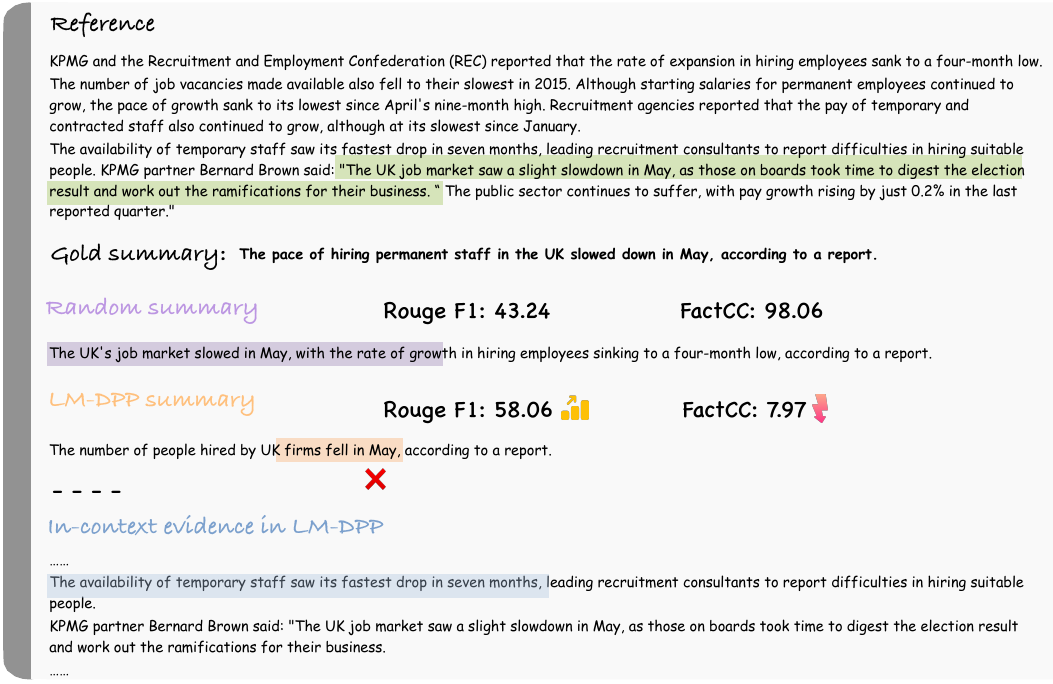}
    \caption{
    Case analysis in XSUM, we compare the performance of Random and LM-DPP on generation quality and fact consistency
    }
    \label{fig:xsum_analysis}
\end{figure*}
\paragraph{Random}
A randomly selected annotation baseline is necessary, as it directly picks unlabeled training instances at random. Ideally, data points selected by any heuristic algorithm should yield better performance compared to it.
\paragraph{Perplexity}
\cite{gonen2022demystifying} reported that lower perplexity correlates with better performance. We rank candidate instances by their perplexity and select the top $|\mathcal{L}|$ instances with the lowest perplexity as our annotation set.
\paragraph{K-means}
As a representative selection method in the series of diversity approaches, we employ clustering techniques. Following \cite{yu2022generate}, we first encode all data points using an \emph{Encoder}, then perform k-means clustering with $|\mathcal{L}|$ clusters and select instances accordingly.
\paragraph{Vote-k}
\cite{su2022selective} selects $|\mathcal{L}|/10$ samples through a graph-based voting mechanism, after which the $|\mathcal{L}|/10$ labeled samples are used as context for the LLMs, to calculate confidence scores for the other unlabeled candidate instances. Finally, the instances are grouped according to percentile ranks of confidence scores, and selection is made through voting within each group.
\paragraph{Fast Vote-k}
A rapid and efficient alternative to Vote-k, it circumvents the use of LLMs to compute confidence scores. It directly selects the $|\mathcal{L}|$ samples with the highest voting scores.

\subsection{Dataset Statistics}
Table \ref{tab:dataset} presents the data statistics of the datasets employed in our experiments. 

\subsection{Prompt Template}
The prompt templates utilized for each task are reported in Table \ref{tab:prompt_template}.



\section{High Uncertainty}
\label{sec:high_uncertainty}
\input{latex/tab/high_perplexity}
Apart from the MNLI and DBpedia datasets, selecting instances of high uncertainty led to a certain degree of performance degradation (Table \ref{tab:high_perplexity}). Therefore, we prioritize the selection of low-uncertainty instances in our experiments and hope to inspire further work in the area of perplexity estimation.

\section{Analysis and supplement}
\subsection{Small Model for scoring}
\label{sec:small_model_score}
\input{latex/tab/small_model_score}
Table \ref{tab:gpt2_score} presents the results of using GPT2 as a surrogate.

\subsection{Fact Consistency in XSUM}
\label{sec:fact_consistency_analysis}
Upon closer analysis (as shown in Figure \ref{fig:xsum_analysis}), we find that in pursuit of diversity and uncertainty in demonstrations, LM-DPP may retrieve content that is topically related but not completely factually consistent. For example, while the source text emphasizes a "The UK job market saw a slight slowdown in May," the LM-DPP generated summary mentions "fell in May," shifting the focal point of the information and potentially misleading readers to interpret a deterioration in actual employment conditions rather than a deceleration in growth rate. This discrepancy is also reflected in the context evidence cited by LM-DPP, which notes "the availability of temporary staff saw its fastest drop in seven months," further reinforcing a negative portrayal of employment circumstances, despite not fully reflecting the source's focus or theme.

We further observe that balancing the Rouge scores with FactCC scores, ensuring factual consistency while maintaining high levels of abstractiveness and textual similarity, presents a significant challenge for LM-DPP. This observation suggests that future research might need to explore more nuanced demonstration selection strategies or introduce stronger fact-checking and correction mechanisms to mitigate the potential risks to factual consistency arising from the pursuit of diversity and uncertainty. This provides valuable insights on how to further optimize the method moving forward.

\subsection{Impact of label coverage}
\label{sec:label_coverage}
\input{latex/tab/mrpc_label_count_example}
\input{latex/tab/trec_label_count_example}
At $\mathcal{L} = 4$, the Acc. of Random and LM-DPP on MRPC and TREC are respectively (47.30, 40.63) and (61.36, 49.64). Combined with Tables \ref{tab:mrpc_label_count_example} and \ref{tab:trec_label_count_example}, it can be seen that as the label coverage increases, performance on MRPC decreases, while TREC shows an expected pattern.
This may be related to the difficulty of the task; moreover, from the perspective of data, an imbalanced label distribution might more closely approximate the statistical characteristics of real-world data.
In certain cases, imbalanced examples could reflect key signals of specific categories, aiding the model in learning effective decision boundaries more swiftly.
We look forward to further research in this area.

\newpage
\input{latex/tab/prompt}

%% file: latex/tab/appendix1.tex
\begin{table}[htbp]
\centering
\resizebox{0.8\linewidth}{!}{
\begin{tabular}{lcc}
\toprule[1.1pt]
              &         \multicolumn{2}{c}{\textbf{QNLI}}        \\ 
                        \cmidrule(lr){2-3} 
        &    $|\mathcal{L}|=16$   &      $|\mathcal{L}|=100$ \\ 
\midrule                                                                                                                                          
$\text{Perplexity}_{\text{avg}}$       & 75.16 & 95.43 \\ 
$\text{Perplexity}_{\text{max}}$       & 143.48 & 278.62 \\

\bottomrule[1.1pt]
\end{tabular}
}
\caption{
Annotation Set (selected by Vote-k) Perplexity Statistics.
}
\label{tab:appendix1}%
\end{table}

%% file: latex/tab/dataset.tex
\begin{table}
\centering
\setlength{\tabcolsep}{4pt}
\def\arraystretch{1.25}
\resizebox{\linewidth}{!}{
\begin{tabular}{l|cc}
\toprule[1.2pt]
\textbf{Dataset}    & \textbf{Task Type} & \textbf{Split}  \\ \hline
SST-5 & Sentiment Classification & 8544/1101/2210 \\
RTE & Natural Language Inference & 2491/277/3000  \\
MNLI & Natural Language Inference & 392702/19647/19643 \\
MRPC & Natural Language Inference & 3668/408/1725  \\
QNLI & Natural Language Inference & 104743/5463/5463 \\
TREC & Topic Classification & 5452/0/500 \\
DBpedia & Topic Classification & 560000/0/70000 \\
Hellaswag & Multiple-choice Question Answering & 39905/10042/10003  \\
COPA & Multiple-choice Question Answering & 1000/0/500 \\
CosmosQA & Multiple-choice Question Answering & 9471/1221/1140 \\
XSUM & Abstractive Summarization & 204045/11332/11334 \\
NQ & Open Domain QA & 307373/7830/0 \\
\bottomrule[1.2pt]
\end{tabular}%
}
\caption{Dataset Statistics in the Experiments.}
\label{tab:dataset}
\end{table}

%% file: latex/tab/high_perplexity.tex
\begin{table}[htbp]
\setlength{\tabcolsep}{4pt}
\def\arraystretch{1.25}
\resizebox{\linewidth}{!}{
\begin{tabular}{lcccccc}
\toprule
\multirow{4}{*}{$\text{LM-DPP}_{\text{high\_uncertainty}}$}   & RTE & MNLI & MRPC & QNLI & SST-5 \\
 & 51.29 & \underline{42.91} & 66.17 & 52.30  & 48.74 & \\  
 
 \cmidrule{2-6} 
 & DBpedia & TREC &HellaSwag & COPA &  \\
 & \underline{93.18} & 81.40 & 66.95 & 83.80 & \\
 \bottomrule
\end{tabular}
}
\caption{Results of selecting high-uncertainty instances (GPTJ + annotation\_size=100+LM-DPP). Improvements in high uncertainty are underlined.}
\label{tab:high_perplexity}
\end{table}

%% file: latex/tab/small_model_score.tex
\begin{table}[htbp]
\setlength{\tabcolsep}{4pt}
\def\arraystretch{1.25}
\resizebox{\linewidth}{!}{
\begin{tabular}{lcccccc}
\toprule

\multirow{4}{*}{$\text{LM-DPP}_{\text{gpt2\_scoring}}$}   & RTE & MNLI & MRPC & QNLI & SST-5 \\
 & 51.96 &41.79 &66.81 &51.43 &47.32 & \\  
 
 \cmidrule{2-6} 
 & DBpedia & TREC &HellaSwag & COPA & Avg  \\
 & 90.67 &81.85 &67.94 &83.09 &64.76 \\
 \bottomrule
\end{tabular}
}
\caption{Results of using GPT2 as a surrogate.}
\label{tab:gpt2_score}
\end{table}

%% file: latex/tab/mrpc_label_count_example.tex
\begin{table}[!t]
\begin{center}
\small
\begin{tabular}{p{0.95\columnwidth}}
\toprule
\textbf{Examples}\\
\midrule
\textbf{LM-DPP}: \makecell[c]{\texttt{equivalent}, \texttt{equivalent}, \\ \texttt{equivalent}, \texttt{equivalent}}\\
\midrule
\textbf{Random}: \makecell[c]{\texttt{equivalent}, \texttt{not equivalent}, \\  \texttt{not equivalent}, \texttt{not equivalent}} \\
\bottomrule
\end{tabular}
\end{center}
\caption{In MRPC, the four demonstration label examples selected by Random and LM-DPP.}
\label{tab:mrpc_label_count_example}
\end{table}

%% file: latex/tab/trec_label_count_example.tex
\begin{table}[!t]
\begin{center}
\small
\begin{tabular}{p{0.95\columnwidth}}
\toprule
\makecell[c]{\textbf{Random}} \\
\midrule

\textbf{Input}: \textit{What are the factors leading to the high teen pregnancy rate in Spartanburg , South Carolina?}\\
\textbf{Label}: \textit{description and abstract concept}\\
\midrule
\textbf{Input}: \textit{Who invented Make-up ?}\\
\textbf{Label}: \textit{\textbf{human being}}\\
\midrule
\textbf{Input}: \textit{Who is the current UN Secretary General ?}\\
\textbf{Label}: \textit{\textbf{human being}}\\
\midrule
\textbf{Input}: \textit{What does God create in the first sentence of the Bible ?}\\
\textbf{Label}: \textit{entity}\\
\midrule
\makecell[c]{\textbf{LM-DPP}} \\
\midrule
\textbf{Input}: \textit{How much caffeine is in a 16 oz cup of coffee ?}\\
\textbf{Label}: \textit{numeric value}\\
\midrule
\textbf{Input}: \textit{What is the fastest growing state in the U.S.A. in 1998 ?}\\
\textbf{Label}: \textit{location}\\
\midrule
\textbf{Input}: \textit{What British female pop singing star of the 1960s and early 1970s was a child actress in the 1940s and '50s}\\
\textbf{Label}: \textit{human being}\\
\midrule
\textbf{Input}: \textit{Why was Muhammad Ali stripped of his title and barred from boxing in 1967 ?}\\
\textbf{Label}: \textit{description and abstract concept}\\

\bottomrule
\end{tabular}
\end{center}
\caption{In TREC, the four demonstration examples selected by Random and LM-DPP.}
\label{tab:trec_label_count_example}
\end{table}

%% file: latex/tab/prompt.tex
\begin{table*}[ht]
\resizebox{\textwidth}{!}{
\centering
\begin{tabular}{lll}
\toprule

\textbf{Dataset} & \textbf{Prompt Template} &\textbf{Example} \\

\midrule
SST-5 & 
\makecell[l]{How do you feel about the following sentence? \\
$\backslash$n \{Input\} $\backslash$n answer:\{Output\}} & 
\makecell[l]{\textbf{Input}: this is a stunning film, a one-of-a-kind tour de force. \\ \textbf{Output}: very positive} \\[12pt]
RTE &
\makecell[l]{\{Input1\}. Based on that information, is the claim \\ \{Input2\} "entailment", or "contradiction"? $\backslash$n answer:\{Output\}} &
\makecell[l]{\textbf{Input1}: No Weapons of Mass Destruction Found in Iraq Yet. \\ \textbf{Input2}: Weapons of Mass Destruction Found in Iraq. \\ \textbf{Output}: contradiction} \\[19pt]
MNLI &
\makecell[l]{\{Input1\}. Based on that information, is the claim \\ \{Input2\} "True", "False", or "Inconclusive"? $\backslash$n answer:\{Output\}} &
\makecell[l]{\textbf{Input1}: Good luck, my friends.\\ 
\textbf{Input2}: I wish my friends luck. \\
\textbf{Output}: True}\\[19pt]
MRPC &
\makecell[l]{Are the following two sentences "equivalent" or "not equivalent"? \\ $\backslash$n \{Input1\}.$\backslash$n \{Input2\}. $\backslash$n answer:\{Output\}} &
\makecell[l]{\textbf{Input1}: Staff writer Dave Michaels contributed to this report. \\ \textbf{Input2}: Staff writers Frank Trejo and Robert Ingrassia contributed to this report. \\ \textbf{Output}: equivalent}
\\[19pt]
BoolQ &
\makecell[l]{\{Input1\}. Based on that information, is the claim \\ \{Input2\} "True", or "False"? $\backslash$n answer:\{Output\}} &
\makecell[l]{\textbf{Input1}: is there going to be another season of Britannia. \\ \textbf{Input2}: In March 2018, is was announced that Sky Atlantic had renewed the show for a second season. \\ \textbf{Output}: True} \\[19pt]
QNLI &
\makecell[l]{\{Input1\}. Based on that information, is the claim \\ \{Input2\} "entailment", or "contradiction"? $\backslash$n answer:\{Output\}} &
\makecell[l]{\textbf{Input1}: About 40,000,000 tons were produced in 1984. \\ \textbf{Input2}: How many tons of bitumen ere produced in 1984? \\ \textbf{Output}: entailment} \\[19pt]
TREC &
content: \{Input\} $\backslash$n \{Output\} &
\makecell[l]{\textbf{Input}: What films featured the character Popeye Doyle ? \\ \textbf{Output}: entity} \\[12pt]
DBpedia &
title: \{Input1\}; content: \{Input2\} $\backslash$n \{Output\} &
\makecell[l]{\textbf{Input1}: Panay Technological College \\ \textbf{Input2}: Panay Technological College is a higher institution in Kalibo Aklan. \\ \textbf{Output}: educational institution} \\[19pt]
Hellaswag &
The topic is \{Input1\}. \{Input2\} $\backslash$n \{Output\} &
\makecell[l]{\textbf{Input1}: Hurling \\ \textbf{Input2}: A group of lacrosse players are shown on a field. they \\ \textbf{Output}: run around, trying to get the ball away from each other.} \\[19pt]
COPA &
\{Input2\}. What was the \{Input1\} of this? $\backslash$n \{Output\} &
\makecell[l]{\textbf{Input1}: cause \\ \textbf{Input2}: My body cast a shadow over the grass. \\ \textbf{Output}: The sun was rising.} \\[19pt]
CosmosQA &
\{Input1\}. \{Input2\} \, $\backslash$n \{Output\} &
\makecell[l]{\textbf{Input1}: El dropped me off at B. 's house. She welcomed El . and me into her home . \\
\textbf{Input2}: Why did she welcome us into the house ? \\ \textbf{Output}: She liked us and enjoys our company .} \\[19pt]
Subj &
Input: \{Input\}. $\backslash$n Type: \{Output\} &
\makecell[l]{\textbf{Input}: katie is a young girl who loves to climb . \\ \textbf{Output}: objective} \\[19pt]
XSUM &
write a short summary:$\backslash$n \{Input\}. $\backslash$n TL;DR: \{Output\} &
\makecell[l]{\textbf{Input}: A lone hiker salutes the aptly named Wet Sleddale Reservoir in Cumbria, as it overflows down a 21 metre high dam wall... \\ \textbf{Output}: Photograph by Jeff Overs / BBC} \\[19pt]
NQ &
Write an answer: \{Input\} $\backslash$n \{Output\} &
\makecell[l]{\textbf{Input}: who is credited with creating the gothic art movement \\ \textbf{Output}: Abbot Suger} \\
\bottomrule
\end{tabular}}

\caption{Prompt templates and corresponding examples used in each dataset.}
\label{tab:prompt_template}
\end{table*}